\documentclass[runningheads]{llncs}
\usepackage[hyphens]{url}
\usepackage{hyperref}
\usepackage[table,xcdraw,dvipsnames]{xcolor}%

\usepackage{eccv}

\usepackage{eccvabbrv}

\usepackage{graphicx}
\usepackage{booktabs}

\usepackage{orcidlink}
\usepackage{xr}

\usepackage[accsupp]{axessibility}  

\usepackage{siunitx}

\usepackage{multirow}

\makeatletter
\newcommand{\printfnsymbol}[1]{%
  \textsuperscript{\@fnsymbol{#1}}%
}
\makeatother

\begin{document}

\title{
    ViG-Bias: Visually Grounded Bias Discovery and Mitigation
}



\author{
    Badr-Eddine Marani\inst{1}\thanks{Equal contribution}
    \and Mohamed Hanini\inst{1}\printfnsymbol{1}
    \and\\ Nihitha Malayarukil\inst{1}
    \and Stergios Christodoulidis\inst{1}
    \and Maria Vakalopoulou\inst{1,3}
    \and Enzo Ferrante\inst{1,2}
}


\institute{
    CentraleSupelec, Université Paris-Saclay, France
    \and Instituto de Ciencias de la Computación, CONICET - Universidad de Buenos Aires, Argentina
    \and Archimedes/Athena RC, Greece
}

\maketitle

\begin{abstract}
\sloppy
The proliferation of machine learning models in critical decision-making processes has underscored the need for bias discovery and mitigation strategies. Identifying the reasons behind a biased system is not straightforward, since in many occasions they are associated with hidden spurious correlations which are not easy to spot. Standard approaches rely on bias audits performed by analyzing model performance in predefined subgroups of data samples, usually characterized by common attributes like gender or ethnicity when it comes to people, or other specific attributes defining semantically coherent groups of images. However, it is not always possible to know a priori the specific attributes defining the failure modes of visual recognition systems. Recent approaches propose to discover these groups by leveraging large vision language models, which enable the extraction of cross-modal embeddings and the generation of textual descriptions to characterize the subgroups where a certain model is underperforming. In this work, we argue that incorporating visual explanations (e.g. heatmaps generated via GradCAM or other approaches) can boost the performance of such bias discovery and mitigation frameworks. To this end, we introduce Visually Grounded Bias Discovery and Mitigation (ViG-Bias), a simple yet effective technique which can be integrated to a variety of existing frameworks to improve both discovery and mitigation performance. Our comprehensive evaluation shows that incorporating visual explanations enhances existing techniques like DOMINO, FACTS and Bias-to-Text, across several challenging datasets, including CelebA, Waterbirds, and NICO++.
The code is available at \url{https://github.com/badrmarani/vig-bias-eccv/}.

    \keywords{
        Unsupervised bias discovery
        \and Bias mitigation
        \and Visual explanations 
    }
\end{abstract}


\section{Introduction}
The increasing adoption of deep learning-based image classification models, notably in systems like facial recognition software, raises concerns related to biases and fairness\cite{buolamwini2018gender}. Biases in such systems can lead to unfair and discriminatory outcomes, making it crucial to identify and address their underlying causes. Unfortunately, determining in which subgroups a given system may exhibit biased behavior is not an easy task. The typical approach to detect biases consists in auditing the system with respect to predefined attributes (often referred to as protected attributes), like gender, sex, age, or ethnicity, by comparing specific fairness metrics \cite{barocas2023fairness}. However, attributes like sex, age, or ethnicity are not always the real causes that explain the poor performance of a given model in a particular set of data samples. For example, in face recognition, a system may systematically fail to classify people with glasses due to the fact that they were not well represented in the training set. In this case, auditing with standard demographic attributes may not be able to reveal the real reason for bias. In other cases, such as the classification of animals or objects, the attributes that characterize the underperforming groups may not be obvious. Spurious correlations could lead to systems that make incorrect predictions because the model has learned these false relationships from the training data \cite{geirhos2020shortcut}, or predictions which are actually correct but for the wrong reasons \cite{sun2023right}. This issue often arises when the model is trained on datasets that do not accurately represent the real-world scenarios on which it is intended to work with, or when the dataset itself contains hidden biases. Thus, discovering the failure modes and providing explanations useful to understand the real reasons behind them becomes a crucial task when designing robust and fair systems. Existing literature refers to this task as \textit{unsupervised discovery of bias} \cite{krishnakumar2021udis}, or \textit{slice discovery} \cite{eyuboglu2022domino}, in the sense that such methods aim at mining the input data for semantically meaningful subgroups (slices) on which the model performs poorly. In this context, a slice is defined as a group of data samples that share a common attribute or characteristic that is not related to the target label.

Recent studies have shown that text explanations \cite{kim2023biastotext} produced by multimodal large vision language models (VLMs) \cite{du2022survey} can help in discovering and describing subpopulations on which a model systematically underperforms, without the need for protected attribute annotations. At the same time, a different line of work \cite{krishnakumar2021udis} has qualitatively demonstrated that visual explanations \cite{alicioglu2022survey} derived from biased models, especially for instances originating from conflicting data slices, predominantly highlight spurious features usually associated with shortcut learning. In this work, we hypothesize that since visual explanations for biased models tend to underscore spurious correlations, they can serve as instrumental tools in informing VLMs when they are used to identify and mitigate undesired biases. To this end, we leverage the descriptive power of recent multimodal VLMs, and combine it with the capabilities of visual explanation mechanisms to uncover biases contributing to systematic failures for unknown slices in visual recognition models. Our extensive empirical evaluation highlights the advantages of the proposed method, being easily adapted to different bias discovery frameworks and systematically boosting their performance.

\noindent \textbf{Contributions:} Under the hypothesis that visual explanations can assist to better identify the reasons behind biased systems, here we introduce Visually Grounded Bias Discovery and Mitigation (ViG-Bias), a simple yet effective strategy which improves the performance of existing bias discovery and mitigation methods based on multimodal embeddings. First, we provide empirical evidence showing that visual explanations of biased models usually focus on spurious correlations. Then, we leverage such explanations to direct the attention of existing bias discovery and mitigation methods towards spurious features, resulting in improved performance in both tasks. We show that ViG-Bias is general enough to be adapted to a wide range of recent methods which address these problems using cross-modal embeddings and textual explanations obtained via VLMs, like DOMINO \cite{eyuboglu2022domino}, FACTS \cite{yenamandra2023facts} and Bias-to-Text \cite{kim2023biastotext}, systematically improving their performance on three challenging datasets. 

\section{Related work}
\label{sec:related_work}

\noindent \textbf{Visual explanations.} Visual explanations in computer vision have largely contributed to the interpretability and trustworthiness of deep learning models. They provide insights into the decision-making process of neural networks by highlighting regions within an input image that are significant for predictions. One prominent technique in this domain is Gradient-weighted Class Activation Mapping (Grad-CAM) \cite{Selvaraju_2019}. Grad-CAM utilizes the gradients flowing into the final convolutional layer of a convolutional neural network to understand which features are most important for a particular decision. By creating a heatmap of these weighted features, Grad-CAM offers a visual representation that demonstrates how the model focuses on different parts of the image to make a decision. This method not only aids in improving model transparency but also helps in diagnosing potential reasons for misclassifications. GradCAM has previously been employed as a description mechanism to interpret failure modes in \cite{krishnakumar2021udis}. However, even though the work of Krishnakumar et al \cite{krishnakumar2021udis} shows the potential of visual explanations for this task, they mostly provide qualitative examples and do not propose an automatic approach to bias discovery and mitigation. Singla and co-workers \cite{singla2022salient} present another interesting work where they generate annotations for visual explanations of the ImageNet dataset. They released a dataset called Salient ImageNet, which provides heatmaps distinguishing between \textit{spurious} and \textit{core} features, where the last ones correspond to a set of visual features that are always a part of the object definition. Even though they employ human supervision to identify the visual explanations related to spurious correlations (making the process difficult to automate), this work shows that such heatmaps can actually be used to pinpoint this type of spurious attributes. Here, we plan to leverage visual explanations (like GradCAM heatmaps and other variants) as a mechanism to direct the attention of bias discovery and mitigation methods towards spurious correlations. \\

\noindent \textbf{Automatic bias discovery using VLMs.}
Several recent efforts have been conducted in the identification of biased data slices using cross-modal embeddings. The DOMINO method \cite{eyuboglu2022domino} combines cross-modal embeddings and error-aware mixture models to find semantically coherent clusters of images where a given classification system is failing. Textual descriptions derived from the cross-modal embeddings are then used to characterize the failure modes that were just discovered. A similar approach is introduced by the FACTS (First Amplify and then Slice to Discover Bias) \cite{yenamandra2023facts} method, where the authors propose to first amplify the sources of bias in the training to ease its identification. This process is carried out in two stages. The first stage involves amplifying the bias by regularizing the model. This is achieved through regularization by specifically increasing the weight decay factor $\lambda$, which penalizes the most significant weights in the model. Regularization is a common technique used to prevent overfitting, but in this context, it serves a dual purpose: it not only helps to prevent overfitting but also forces the model to rely more on the spurious attribute by simplifying the hypothesis space. The second stage consists of identifying underperforming data segments that exhibit unique correlations by employing mixture modeling within a feature space aligned with bias, a method known as correlation-aware slicing, similar to the aforementioned DOMINO. In \cite{kim2023biastotext}, a different approach called Bias-to-Text (B2T) was introduced. B2T uses linguistic interpretation to identify and mitigate biases in vision models, including image classifiers. By generating linguistic descriptions of images using VLMs, B2T extracts keywords indicative of bias from the wrongly predicted examples, enabling a more focused understanding and correction of prejudices within the models. 

In this work, we will show how visual explanations can boost the performance of these three approaches, by helping to better focus the model's attention on areas containing spurious features.\\

\noindent \textbf{Bias mitigation: improving robustness to spurious correlations.} Discovering conflicting slices is important, but mitigating such biases is also crucial. A common approach is based on group distributionally robust optimization (GroupDRO) \cite{sagawa2020distributionally}. GroupDRO aims to enhance model performance across different groups or subpopulations in the data, based on predefined protected attributes like race or gender. The objective is to minimize the worst-case loss across these groups, thereby encouraging fairness and reducing disparities in model performance among them. However, one notable drawback of this technique is the requirement to assign specific group labels to every dataset entry, which can lead to substantial increases in annotation costs, and requires to know before-hand such groups. Another existing approach in the bias mitigation field is JTT (Just Train Twice) \cite{liu2021just}. JTT starts with the standard training of the model to pinpoint examples that have been misclassified. Following this, the training process is adjusted by reweighting the dataset, giving greater importance to these previously misclassified examples. JTT forces the model to pay more attention to examples where the model is failing, thereby enhancing its performance on those specific examples. Here, we will employ JTT as a baseline approach for comparison, and propose ways to integrate visual explanations into the group definitions for GroupDRO.

\section{Preliminaries and Problem Formulation}

Following the formulation introduced in \cite{yenamandra2023facts}, let $\mathcal{X}$ and $\mathcal{Y}$ be the input (images) and output (label) spaces, respectively, and $\mathcal{D}=\{(x_i, y_i)\}_{i=1}^N$ be a training dataset of $N$ samples drawn from $\mathcal{X} \times \mathcal{Y}$. We are interested in learning a classification model $f_\theta: \mathcal{X} \rightarrow \mathcal{Y}$, parameterized by $\theta \in \Theta$, by minimizing the average loss (\eg cross-entropy) across training samples \ie $\frac{1}{N} \sum_{i=1}^N \ell(f_\theta(x_i), y_i)$. We work under the hypothesis that, due to the presence of spurious correlations, the model could incur in biased predictions, failing to generalize at test time. To formulate the spurious correlations, let $\mathcal{A}=\{a_1, \ldots, a_n\}$ denote a set of spurious attributes, where $a_i(x) \in \{0,1\}$ indicates that the attribute $a_i$ is present in the image $x$. In the presence of spurious correlation between an attribute $a_i$ and a target label $y$, the model learns to rely on it rather than the real target features to make predictions. This results in the model performing poorly on examples of class $y$ that do not contain the spurious feature $a_i$.

The dataset $\mathcal{D}$ is partitioned into different and not equally distributed groups or slices. A group is defined based on the combination of the target labels $y \in \mathcal{Y}$ and the spurious attributes $a_i \in \mathcal{A}$ that spuriously correlates with the label (\ie $\mathcal{G} = \mathcal{A} \times \mathcal{Y}$). Formally, given a pair $(a_i, y^\star)$, we define a group $g(a_i, y^\star)$ as the set of data samples labeled with $y^\star$ which contain attribute $a_i$, i.e. where $a_i(x) = 1$.

\begin{figure}[tb]
  \centering
  \begin{subfigure}[c]{0.35\textwidth}
    \includegraphics[width=\textwidth]{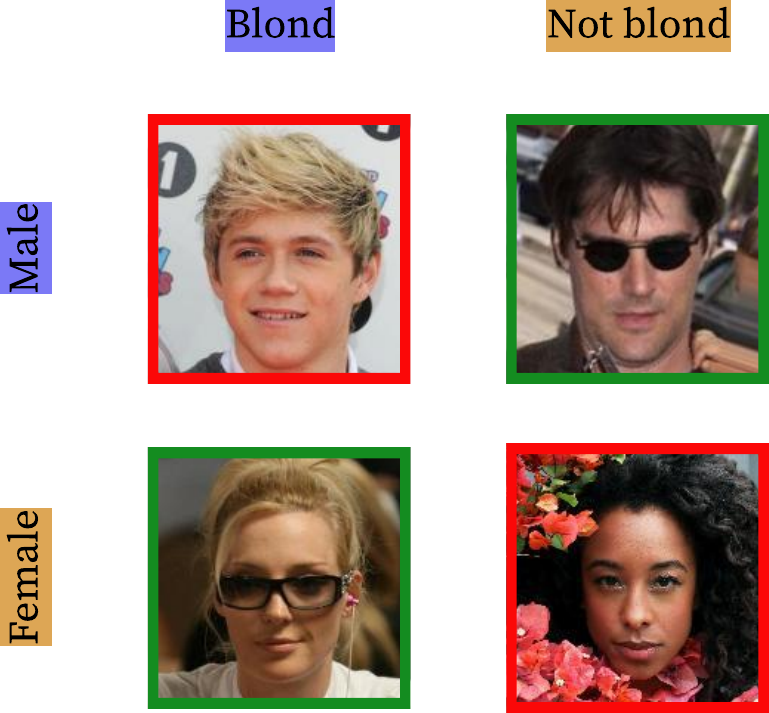}
    \caption{In the CelebA dataset, the label, \textit{hair color} is spuriously correlated with the attribute \textit{gender}.}
    \label{fig:celeba_groups-a}
  \end{subfigure}%
  ~~~~
  \begin{subfigure}[c]{0.35\textwidth}
    \includegraphics[width=\textwidth]{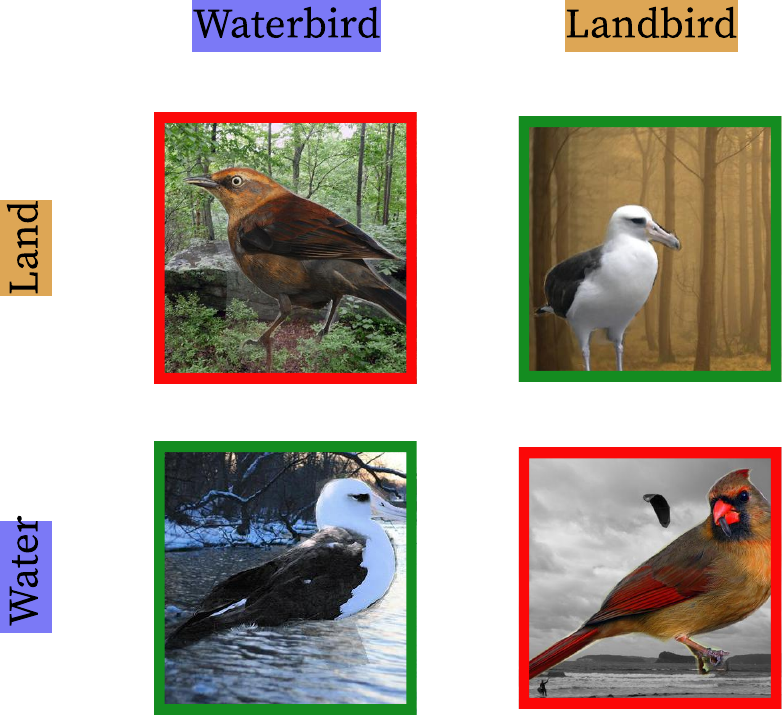}
    \caption{In the Waterbirds dataset, the label, \textit{bird type} is spuriously correlated with the attribute \textit{background}.}
    \label{fig:waterbirds_groups-b}
  \end{subfigure}
  \caption{The four groups in the CelebA (left) and Waterbirds (right) datasets are based on combinations of the spurious attribute and the label. The groups highlighted in green have the most training samples, whereas the groups highlighted in red have the least training samples.}
  \label{fig:celeba_waterbirds_groups}
\end{figure}

For example, in the CelebA dataset\cite{liu2015faceattributes}, where the task involves classifying people with blond hair, the spurious attribute is the gender, as there is a higher frequency of blond women in the dataset. \Cref{fig:celeba_waterbirds_groups} illustrates this issue, and a similar case found in the Waterbirds dataset \cite{sagawa2020distributionally}, highlighting the classes with the least training samples per group.
The groups with the largest number of samples are where the correlations hold, whereas the groups with the smallest number of samples are where the correlations do not hold.
Training a model by simply minimizing the average error can lead to rely on spurious features, like gender to predict hair color in the CelebA dataset. Consequently, the model may perform poorly on minority groups where this spurious correlation does not hold. Following \cite{yenamandra2023facts}, we consider a function $M: \mathcal{A} \rightarrow \mathcal{Y}$ that maps each spurious attribute to the unique label that is most associated with it. Then, the unsupervised bias discovery problem is defined as finding the groups (or slices) $G$, where attribute $a_i$ is spuriously correlated with label $y$ as:
\begin{equation}
    G =
    \Big\{
        g(a_i, y) \,|\, \forall a, y \in \mathcal{A} \times \mathcal{Y}, \; M(a) = y
    \Big\}
\end{equation}

\section{ViG-Bias: Visually Grounded Bias Discovery and Mitigation}
\label{sec:vig-bias}

In this section, we present the proposed method for discovering biases using visual explanations. We first provide quantitative evidence showing that visual explanations tend to correlate with spurious features, and then propose using them to drive the attention and improve performance for a variety of bias discovery and mitigation methods.

Consider a dataset $D$ with samples $(x, y, a) \in \mathcal{D}$ drawn from $\mathcal{X} \times \mathcal{Y} \times \mathcal{A}$, where $a$ is spuriously correlated with the label $y$. We are interested in discovering the visual features upon which a model has relied to make a prediction. Prior work \cite{joshi2023mitigating, singla2022salient} suggests that, under the presence of spurious correlations, visual explanations for models trained with standard empirical risk minimization (ERM) tend to focus on such shortcuts, while ignoring the core features. We thus resort to GradCAM heatmaps generated using an image classifier that was trained with standard ERM on the dataset of interest $\mathcal{D}$. Such heatmaps will then be used as visual explanations to guide the attention of bias discovery methods (alternative visual explanation methods like ScoreCAM \cite{wang2020scorecam}, FullGrad \cite{srinivas2019fullgradient},
GradCAM++ \cite{Chattopadhay_2018} are also evaluated in \Cref{ablations}).
GradCAM heatmaps are obtained by first computing the gradients of the target class with respect to the feature maps of the last convolutional layer, as neurons in this layer offer best compromise between high-level semantics and detailed spatial
information \cite{Selvaraju_2019}. These gradients flowing back are then globally-averaged over the width and height dimensions to obtain the  importance weights of every feature map in this last layer. Then, the ReLU activation function is applied on the weighted average of the feature maps, so that only the features that have a positive influence on the class of interest are finally highlighted. These heatmaps are finally normalized between 0 and 1.

\subsection{Are Visual Explanations Focusing on Spurious Correlations?}
\label{sec:visual_explan_sp_corr}
The work of Krishnakumar \cite{krishnakumar2021udis} and colleagues provides qualitative examples showing that GradCAM heatmaps for biased models tend to focus on spurious correlations. Here we perform a more systematic quantitative evaluation to confirm this hyphotesis. To this end, we design a simple experiment using CelebA (where the conflicting slices are well known) that consists of three steps: 

 \begin{enumerate}
     \item We first construct binary segmentation masks $m^c_i$ and $m^s_i$ for all images $x_i$ in both datasets, segmenting the core features associated with the object of interests ($m^c_i$ representing, for example, the \textit{hair} in CelebA), and the spurious attribute ($m^s_i$ representing the \textit{face}) as shown in \Cref{fig:celeba_waterbirds_groups}. We do so by using LangSAM \footnote{\url{https://github.com/luca-medeiros/lang-segment-anything}}, an open-source model that combines Segment Anything Model (SAM) \cite{kirillov2023segment} with  GroundingDINO \cite{liu2023grounding} to enable instance segmentation from text prompts.  
     
    \item We then proceed to create binary masks $b_i$ by thresholding the heatmaps obtained via GradCAM as $b_i = \mathrm{GradCAM}(x_i) \geq \tau$. These masks represent the area where the biased model focuses its attention (we use $\tau=0.7$ in our experiment which was choosen by visual inspection).

    \item Finally, we measure the intersection over union (IoU) between the GradCAM masks $b_i$, and both, the core ($m^c_i$) and spurious ($m^s_i$) segmentations.
\end{enumerate}

\begin{figure}[tb]
    \centering
    \includegraphics[width=\textwidth]{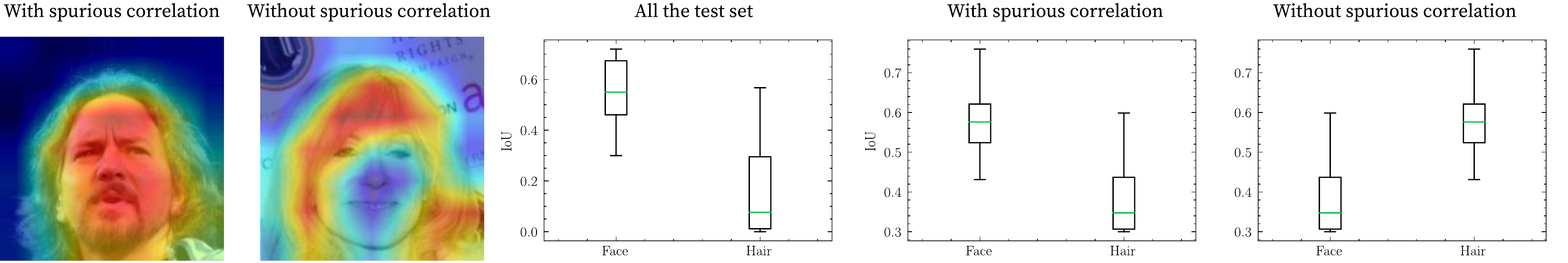}
    \caption{Visual explanation methods (\eg GradCAM\cite{Selvaraju_2019} for the spurious attribute \textit{blond}) help with identifying spurious correlations. We use the Intersection-over-Union (IoU) metric to measure the percentage of overlap between a binary image representing the spurious feature and the image we get after applying a visual explanation method.}
    \label{fig:boxplot}
\end{figure}

\Cref{fig:boxplot} shows the results of this motivational experiment. As it can be observed, GradCAM masks coming from bias-conflicted cases (with spurious correlations) tend to present a significantly higher overlap with the spurious features, what in this case implies focusing directly on the face instead of the hair. With this quantitative experiment confirming our initial hypothesis, we proceed to describe the ViG-Bias framework, showing how visual explanations can be easily integrated to improve performance in a variety of existing bias discovery and mitigation methods. 

\subsection{Improving Unsupervised Bias Discovery via Visual Explanations}
\label{sec:improving_bias_discovery}
ViG-Bias is based on a simple yet effective idea: use visual explanations to direct the attention of bias discovery methods towards real spurious features. Let us define a mapping function $h: \mathcal{X} \rightarrow \mathcal{X}$, that is applied to an input image $x$ to highlight areas where a given classifier is focusing, as:
\begin{equation}
    \label{eq:h}
    h(x)=\bbbone\Big\{x \odot\,\mathrm{GradCAM}(x) \geq \tau\Big\}
\end{equation}
where $\odot$ represents the Hadamard (element-wise) product of the input image $x$ and the heatmap produced by the GradCAM (or other visual explanation method). We argue that such simple function can help to improve the performance of recent bias discovery methods based on cross-modal feature clustering (e.g. DOMINO \cite{eyuboglu2022domino} and FACTS \cite{yenamandra2023facts}) and image captioning (e.g. Bias-to-text \cite{kim2023biastotext}).
In the following sections, we discuss how visual explanations can be incorporated into each of the aforementioned methods.

\noindent\textbf{Visually Grounded DOMINO (ViG-DOMINO):} DOMINO uses cross-modal embeddings to identify coherent conflicting slices and generate natural language descriptions to describe them. To this end, they first project the images into a cross-modal embedding (e.g CLIP \cite{radford2021learning}). Then, given a classifier that was trained using a standard ERM procedure, predictions for each image are obtained. These predictions are then combined with the cross-modal embeddings and the true labels associated with the input image. An error-aware mixture model is finally fit to cluster slices that are homogeneous not only semantically, but also with respect to error type (e.g. all false positives). Finally, natural language descriptions of the discovered slices are produced to describe characteristics shared between examples in the discovered slices. For more details about DOMINO please refer to \cite{eyuboglu2022domino}.

We modify DOMINO by simply pre-processing the images before the cross-modal embeddings are obtained, i.e. we generate heatmaps using the pretrained classifier and extract embeddings for $h(x)$ instead of $x$. Note that other types of visual explanation methods could also be used, as discussed in \Cref{ablations}.

\noindent\textbf{Visually Grounded FACTS (ViG-FACTS):} The FACTS method is similar to DOMINO, but it consists of a two-stages process, where biases are first amplified and then slicing is performed. In the first stage, we increase the model's reliance on spurious, often bias-indicative correlations, which are amplified using regularization. In general, regularization helps in preventing overfitting and encourages the model to learn more robust patterns that are assumed to be applicable beyond the training set. However, in FACTS, the regularization is applied to constrain the model's capacity and learn the bias-aligned slices where spurious correlation holds. This leads to the model developing a strong dependency on these correlations, making the biases more pronounced and identifiable. Once these biases are amplified and more clearly delineated, we proceed to the second stage -\textit{slicing}- where an error-aware mixture model that uses the CLIP embeddings similarly to DOMINO is applied.

We modify FACTS the same way as we modified DOMINO, by applying the mapping $h(x)$ to the input images before embedding them to CLIP.

\begin{figure}[tb]
    \centering
    \includegraphics[width=\textwidth]{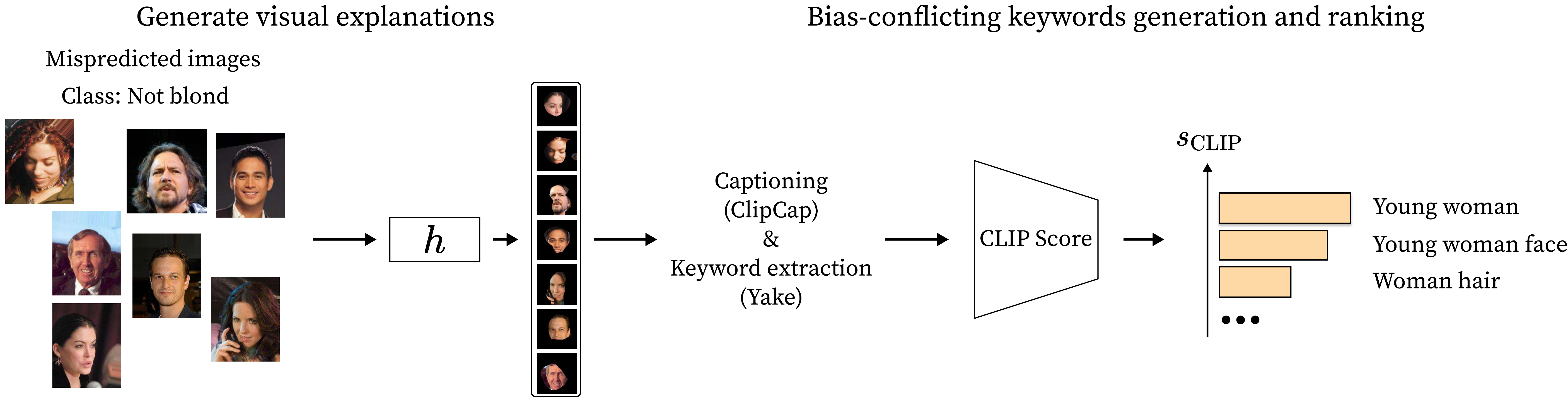}
    \caption{The B2T framework treats visual biases as language, allowing to \emph{(1)} uncover new biases by interpreting keywords and \emph{(2)} mitigate biases in models by leveraging these identified keywords. Through B2T, spurious correlations between attributes like \textit{female} and \textit{blond} are uncovered. To enhance this process, we suggest integrating a visual explanation component before keyword generation. This addition aims to improve the relevance of identified keywords and assess the effectiveness of using these keywords in debiasing.}
    \label{fig:b2t_method_with_visual_explanations}
\end{figure}

\noindent\textbf{Visually Grounded Bias-to-Text (ViG-B2T):}
In contrast to the methods described above that focus on discovering conflicting slices on the cross-modal embeddings, B2T first generates textual image descriptions which are then processed to identify the conflicting slices. The ClipCap \cite{mokady2021clipcap} image captioning model is used to generate textual descriptions. Then, keywords are extracted for the uncorrectly predicted samples, and 
ranked using the CLIP score to determine which ones are related to the spurious attributes, based on their similarity with correctly or incorrectly classified images. Formally, the CLIP score is given by:
\begin{equation}
    s_{\mathrm{CLIP}}(k, \mathcal{D}) =
    \mathrm{sim}(k, \mathcal{D}_{\mathrm{wrong}}) -
    \mathrm{sim}(k, \mathcal{D}_{\mathrm{correct}})
\end{equation}
where $\mathcal{D}_{\mathrm{wrong}}$ and $\mathcal{D}_{\mathrm{correct}}$ are the set of correctly and incorrectly classified images, respectively, and $\mathrm{sim}(a, \mathcal{D})$ is the similarity between the keyword $k$ and the dataset $\mathcal{D}$. Keywords with the highest CLIP score are considered to describe the spurious attributes. \Cref{fig:b2t_method_with_visual_explanations} illustrates the overall framework of B2T method.

We modify B2T and improve the quality of the bias description keywords by focusing the attention of the image captioning model into the spurious features using visual explanations. Therefore, before generating the captions, we first apply function $h(x)$ to the images, then generate the captions, extract the keywords and rank them as originally proposed in\cite{kim2023biastotext}. \Cref{fig:b2t_method_with_visual_explanations} illustrates the B2T method, and how we utilize visual explanations.

\begin{figure}[tb]
    \centering
    \includegraphics[width=\textwidth]{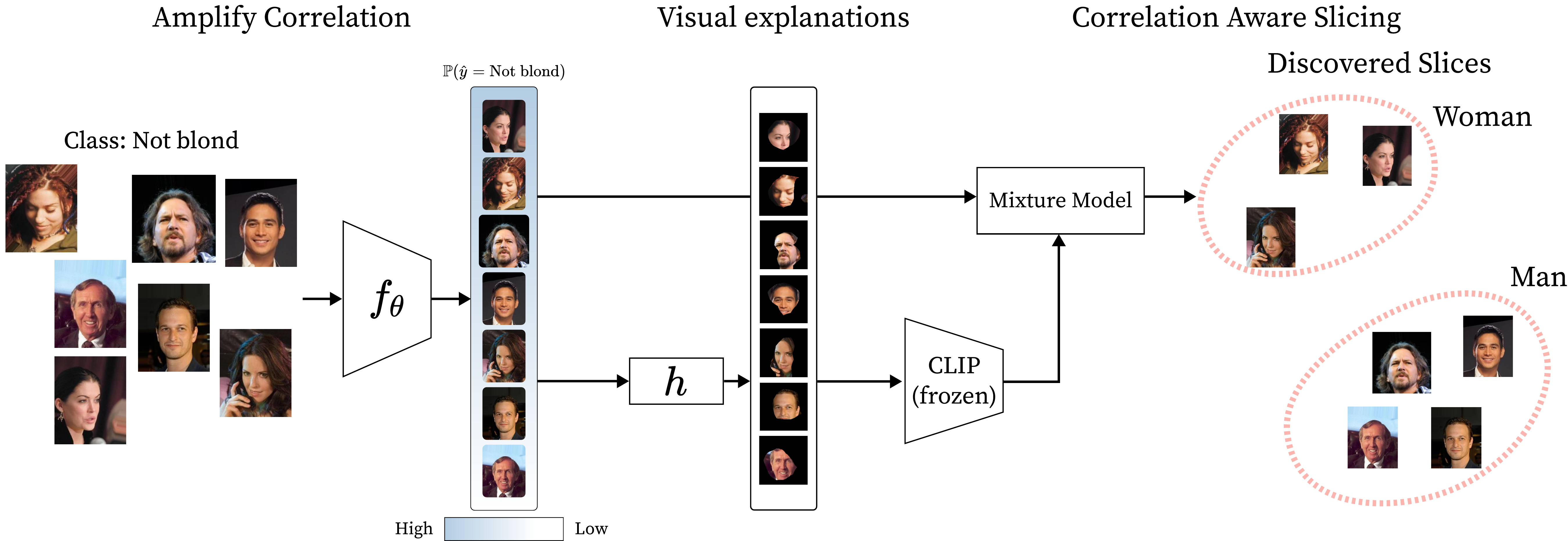}
    \caption{Our objective is to pinpoint slices of data where a spurious correlation exists between a task-irrelevant attribute (such as gender) and the label of interest. For instance, in the given example, women who are not blond correspond to a bias-conflicting slice, while not blond men form a bias-aligned slice. FACTS amplifies correlations, aiming to establish a straightforward bias-aligned hypothesis. Then, it applies a visual explanation to the dataset using the amplified model. Finally, correlation-aware slicing is executed, a process in which clustering is conducted within the bias-amplified feature space. To enhance this process, we suggest integrating a visual explanation component $h$ before keyword generation. 
    }
    \label{fig:facts_method_with_visual_explanations}
\end{figure}




\subsection{Improving bias mitigation via visual explanations}
\label{sec:improving_bias_mitigation}

Although the methods described above are effective in discovering biases in image classifiers, they do not provide a direct way to mitigate them. In this section, we describe how bias mitigation methods can benefit from visual explanations to improve worst-group accuracy. Here we focus on two alternatives: using language-based zero-shot classifiers, and Group Distributionally Robust Optimization (GroupDRO)\cite{sagawa2020distributionally}.

\noindent\textbf{Zero-shot Classifier.}
Language-based Zero-shot (ZS) classification can be defined as the task of predicting a class that has not been explicitly seen during training, by levereging the semantic understanding encoded in its pre-trained embeddings (\eg, using cross-modal embeddings like CLIP \cite{radford2021learning}). Prior studies proposed several prompting strategies to mitigate biases in ZS classifiers. \cite{zhang2022contrastive} employs group-informed prompting, which performs zero-shot
classification using prompts with group information (e.g., “a waterbird on a land background”), what has shown to reduce the gap between worst and average-class accuracy. Since this approach still requires having access to the group names, which is not the case in our setting, we follow \cite{kim2023biastotext} and employ a prompting strategy that mitigates biases in ZS classifiers by augmenting the base prompt with conflicting keywords discovered by B2T and our visually grounded variant, ViG-B2T. Importantly, we show that the words discovered by ViG-B2T actually lead to improved mitigation performance.

\noindent\textbf{Group Distributionally Robust Optimization (GroupDRO).}
As discussed in the \Cref{sec:related_work}, GroupDRO\cite{sagawa2020distributionally} is effective at mitigating biases but it requires pre-defined group annotations.
Formally, for some loss functions $\ell$ (\eg cross-entropy) and training data $\{(x_i, y_i, g_i)\}_{i=1}^{N}$ with $N$ samples, the GroupDRO objective is given by:
\begin{equation}
    \min_{\theta \in \Theta}
    \;\Bigg\{
        \max_{g \in \mathcal{G}} \frac{1}{n_g} \sum_{i=1 | g_i=g}^{n_g} \ell(f_\theta(x_i), y_i)
    \Bigg\}
\end{equation}
where $n_g$ is the number of samples per group. 
The original B2T proposes inferring group annotations using the language-based zero-shot classifier, CLIP\cite{radford2021learning}, and then re-train a classifier using the GroupDRO algorithm with the new group annotations. Here, we evaluate the performance of GroupDRO when trained using the keywords inferred by the original B2T and our proposed ViG-B2T.

\section{Experiments}

We study three datasets: CelebA \cite{liu2015faceattributes}, Waterbirds \cite{sagawa2020distributionally} and NICO++ \cite{zhang2022nico}, which are standard benchmarks for evaluating the discovery and mitigation of biases caused by spurious correlations. All the details about the datasets can be found in Supplementary Material A. Moreover, all implementation details of our method are presented in Supplementary Material B.

\subsection{Evaluation Protocol}
\noindent\textbf{Bias Discovery Metrics.}
In our study, we adopt the \texttt{Precision@k} metric \cite{eyuboglu2022domino} to assess the effectiveness of our slice discovery methodology in identifying bias-conflicting slices within a dataset. This metric measures the proportion of the top $k$ elements in the discovered slice that are in the ground-truth slice. For more details we refer to \cite{eyuboglu2022domino}.

\noindent\textbf{Bias Mitigation Metrics.} To evaluate the bias mitigation performance, we use two different metrics: average and worst-case accuracy. Following\cite{sagawa2020distributionally}, we report an adjusted average accuracy, which is the test accuracy across all groups, weighted by the number of samples in each group in the training data. The worst-group accuracy (WGA) is the lowest accuracy across all groups defined as,
    $\min_{g \in \mathcal{G}} \frac{1}{n_g} \sum_{i=1 | g_i=g}^{n_g} \bbbone\Big\{f_\theta(x) = y\Big\}$
    where $n_g$ is the number of samples per group. 
    
\subsection{Comparison Baselines}
\noindent\textbf{Bias Discovery Baselines.}
Our objective is to assess the impact of incorporating visual explanations into a bias discovery method and determine whether it enhances the effectiveness of the bias discovery process. Thus, the baseline methods consist of the standard versions of the bias discovery methods, for this comparison we include B2T \cite{kim2023biastotext}, FACTS \cite{yenamandra2023facts}, and DOMINO \cite{eyuboglu2022domino}. 

\noindent\textbf{Bias Mitigation Baselines.} Regarding debiasing methods, our aim is to assess the influence of visual explanations on their formulation. As outlined in \Cref{sec:improving_bias_mitigation}, our approach involves extracting keywords associated with spurious attributes and informed using visual explanations, forming groups through zero-shot classification based on these keywords and labels, and subsequently training a GroupDRO\cite{sagawa2020distributionally} model based on the generated groups. We intend to compare these methods with the standard versions which do not incorporate visual explanations. Additionally, we will compare them with a standard version of GroupDRO\cite{sagawa2020distributionally} where groups are built using the ground-truth attributes already present in the dataset, as well as with another mitigation method, JTT \cite{liu2021just}, which does not rely on group information.



\vspace{-3mm}
\section{Results}

\subsection{Evaluating Bias Discovery} 
We first evaluate the quality of the predicted slices with and without utilizing visual explanations in \Cref{tab:precision}.
As in \cite{yenamandra2023facts}, we use the metric \texttt{Precision@k}, and we set $k=10$. Unlike DOMINO and FACTS, B2T does not assume that each target label may be mapped to multiple spurious features. Moreover, the discovered keywords in B2T reflect only the dominant spurious feature per class. That is why we could not report its results on the NICO++ dataset (note that the original B2T publication does not report results on NICO++ neither). Nonetheless, the accuracy of the inferred slices when using visual explanations in B2T on Waterbirds and CelebA datasets shows systematic improvement. Importantly, incorporating visual explanations into bias discovery methods yields either the same or better \texttt{Precision@k} across all five datasets.

\begin{table}[tb]
    \caption{Visual explanations enhance the effectiveness of bias discovery methods. We report $precision@k$ as a metric to assess the quality of slices generated by each method.}
    \centering
    \resizebox{0.8\textwidth}{!}{
    \begin{tabular}{
        l
        c@{\hskip 0.1in}
        c@{\hskip 0.1in}
        c@{\hskip 0.1in}
        c@{\hskip 0.1in}
        c
    }
        \toprule
        \multirow{2}{*}{Method} & \multicolumn{5}{c}{Datasets} \\
        \cmidrule(lr){2-6}
               &  Waterbirds & CelebA & NICO++$^{75}$ & NICO++$^{90}$ & NICO++$^{95}$\\
        \midrule
        DOMINO\cite{eyuboglu2022domino} & $90.0\%$ & $87.0\%$ & $24.0\%$ & $24.0\%$ & $24.0\%$ \\
        ViG-DOMINO (ours)  & $\mathbf{92.0\%}$ & $\mathbf{90.0\%}$ & $\mathbf{25.0\%}$ & $\mathbf{24.0\%}$ & $\mathbf{25.4\%}$ \\
        \midrule
        B2T\cite{kim2023biastotext}  & $92.0\%$ & $64.0\%$ & - & - & -\\
        ViG-B2T (ours) & $\mathbf{97.0\%}$ & $\mathbf{70.0\%}$ & - & - & - \\
        \midrule
        FACTS\cite{yenamandra2023facts} & $100.0\%$ & $100.0\%$ & $55.0\%$ & $60.8\%$ & $61.0\%$ \\
        ViG-FACTS (ours)    & $\mathbf{100.0\%}$ & $\mathbf{100.0\%}$ & $\mathbf{60.0\%}$ & $\mathbf{66.7\%}$ & $\mathbf{65.0\%}$ \\
        \bottomrule
    \end{tabular}
    }
    \label{tab:precision}
\end{table}

\subsection{Evaluating Bias Mitigation}
We then assess the effectiveness of identified slices and keywords in reducing spurious correlations. Initially, we  assess the average and worst-group accuracies for zero-shot classification using various prompting strategies. Then, we compare debiasing models using the GroupDRO algorithm, each using the discovered keywords to infer the new group annotations across all datasets.

\sloppy
\noindent\textbf{Evaluating Zero-Shot Classification.}
Results shown in \Cref{tab:zs} reveal that incorporating visual explanations enhances worst group accuracy across all datasets and reduces the gap between worst and average accuracy, resulting in significant improvements. For instance, in the Waterbird dataset, we achieve an 8\% increase in worst group accuracy compared to Zero-shot classification without visual explanation. Similarly, in CelebA, we observe a 1.2\% enhancement in worst group accuracy.
\begin{table}[tb]
    \caption{Visual explanations improve the effectiveness of bias mitigation approaches. We report worst group accuracy and average accuracy across all groups (the higher the better) and gap between these two (the lower, the better).}
    \centering
    \resizebox{\textwidth}{!}{
    \begin{tabular}{
        l
        ccc
        ccc
        ccc
        ccc
        ccc
    }
        \toprule
            Prompting strategy &
            \multicolumn{3}{c}{Waterbirds} &
            \multicolumn{3}{c}{CelebA} &
            \multicolumn{3}{c}{NICO++$^{75}$} &
            \multicolumn{3}{c}{NICO++$^{90}$} &
            \multicolumn{3}{c}{NICO++$^{95}$}\\
        \cmidrule(lr){2-4}
        \cmidrule(lr){5-7}
        \cmidrule(lr){8-10}
        \cmidrule(lr){11-13}
        \cmidrule(lr){14-16}
        & Worst & Avg. & {\cellcolor[HTML]{96FFFB}Gap}
        & Worst & Avg. & {\cellcolor[HTML]{96FFFB}Gap}
        & Worst & Avg. & {\cellcolor[HTML]{96FFFB}Gap}
        & Worst & Avg. & {\cellcolor[HTML]{96FFFB}Gap}
        & Worst & Avg. & {\cellcolor[HTML]{96FFFB}Gap}\\
        \midrule

        ZS w/ Base prompt\cite{radford2021learning}& 
             $51.0\%$ & $79.0\%$ & {\cellcolor[HTML]{96FFFB} $28\%$} & $69.4\%$ & $81.9\%$ &{\cellcolor[HTML]{96FFFB}$12.5\%$}& $70.7\%$ & $76.0\%$ & {\cellcolor[HTML]{96FFFB} $5.3\%$} & $70.3\%$ & $76.6\%$ & {\cellcolor[HTML]{96FFFB} $6.3\%$} & $68.2\%$ & $77.3\%$ & {\cellcolor[HTML]{96FFFB} $9.1\%$}\\
        ZS + Group labels\cite{zhang2022contrastive}&
             $50.3\%$ & $\mathbf{82.7\%}$ & {\cellcolor[HTML]{96FFFB} $32.4\%$} & $71.6\%$ & $\mathbf{90.2\%}$ & {\cellcolor[HTML]{96FFFB} $18.6\%$} & $75.3\%$ & $76.8\%$ & {\cellcolor[HTML]{96FFFB} $1.5\%$} & $75.8\%$ & $77.2\%$ & {\cellcolor[HTML]{96FFFB} \textbf{$\mathbf{1.4\%}$}} & $75.1\%$ & $77.7\%$ & {\cellcolor[HTML]{96FFFB} $2.6\%$}\\
        ZS + B2T Groups\cite{kim2023biastotext}&
             $55.0\%$ & $76.3\%$ & {\cellcolor[HTML]{96FFFB} $21.3\%$} & $77.5\%$ & $86.4\%$ & {\cellcolor[HTML]{96FFFB} $8.9\%$} & $77.0\%$ & $77.6\%$ & {\cellcolor[HTML]{96FFFB} \textbf{$\mathbf{0.6\%}$}} & $69.9\%$ & $75.0\%$ & {\cellcolor[HTML]{96FFFB} $5.1\%$}& $77.1\%$ & $75.0\%$ & {\cellcolor[HTML]{96FFFB} $\mathbf{2.1\%}$}\\
        ZS + ViG-B2T Groups (ours)&
             $\mathbf{63.1\%}$ & $77.8\%$ & {\cellcolor[HTML]{96FFFB} $\mathbf{14.7\%}$}& $\mathbf{78.2\%}$ & $85.2\%$ &{\cellcolor[HTML]{96FFFB} $\mathbf{7\%}$}& $\mathbf{77.9\%}$ & $\mathbf{79.4\%}$ &{\cellcolor[HTML]{96FFFB} $1.5\%$}& $75.3\%$ & $\mathbf{80.6\%}$ &{\cellcolor[HTML]{96FFFB} $5.3\%$}& $74.6\%$ & $\mathbf{81.1\%}$ &{\cellcolor[HTML]{96FFFB} $6.5\%$}\\
        \bottomrule
    \end{tabular}
    }
    \label{tab:zs}
\end{table}

\noindent\textbf{Debiasing Classifiers With GroupDRO.}
We compare the bias mitigation performance when training a classifier using GroupDRO considering three different group definitions:\emph{(1)}  using  the ground-truth attributes (e.g. \textit{male} and \textit{female} for CelebA, or \textit{land} and \textit{water} for Waterbirds), \emph{(2)} using inferred groups obtained from standard B2T \cite{kim2023biastotext} and \emph{(3)} incorporating visual explanations in the creation of the inferred groups via ViG-B2T. Additionally, we compare our approach with JTT \cite{liu2021just}, which do not use group information, and standard ERM. Results are presented in \Cref{tab:wga}, showing that integrating visual explanations enhances worst group accuracy across all datasets, and yields the lowest gap between worst and average accuracy. 
\begin{table}[tb]
    \caption{Incorporating visual explanations improves the effectiveness of bias mitigation approaches, particularly in terms of worst group accuracy, a commonly utilized metric in bias mitigation tasks.}
    \centering
    \resizebox{\textwidth}{!}{
    \begin{tabular}{
        l
        ccc
        ccc
        ccc
    }
        \toprule
            Method &
            \multicolumn{3}{c}{Uses group information} &
            \multicolumn{3}{c}{Waterbirds} &
            \multicolumn{3}{c}{CelebA}\\
        \cmidrule(lr){5-7}
        \cmidrule(lr){8-10}
        &
        &&
        & Worst & Avg. & {\cellcolor[HTML]{96FFFB} Gap.}
        & Worst & Avg. & {\cellcolor[HTML]{96FFFB} Gap.}\\
        \midrule

        ERM
            && &\multirow{2}{*}{No}& $62.4\%$ & $97.7\%$ & {\cellcolor[HTML]{96FFFB} $35.3\%$} & $47.0\%$ & $94.9\%$ & {\cellcolor[HTML]{96FFFB} $47.9\%$}\\
        JTT\cite{liu2021just}
            && && $86.3\%$ & $92.8\%$ & {\cellcolor[HTML]{96FFFB} $6.5\%$} & $82.0\%$ & $88.0\%$ & {\cellcolor[HTML]{96FFFB} $6\%$} \\
        \midrule
        GroupDRO w/ original groups\cite{sagawa2020distributionally}
            && &\multirow{3}{*}{Yes}& $88.0\%$ & $93.7\%$ & {\cellcolor[HTML]{96FFFB} $5.7\%$} & $88.4\%$ & $91.6\%$ & {\cellcolor[HTML]{96FFFB}$3.2\%$}\\
        GroupDRO w/ B2T groups \cite{kim2023biastotext}
            && && $88.3\%$ & $93.8\%$ & {\cellcolor[HTML]{96FFFB}$5.5\%$} & $87.8\%$ & $92.7\%$ & {\cellcolor[HTML]{96FFFB}$4.9\%$}\\
        GroupDRO w/ ViG-B2T groups (ours)
            && && $\mathbf{90.2\%}$ & $93.4\%$ & {\cellcolor[HTML]{96FFFB} $\mathbf{3.2\%}$} & $\mathbf{91.0\%}$ & $93.9\&$ & {\cellcolor[HTML]{96FFFB} $\mathbf{2.9\%}$}\\
       \bottomrule
    \end{tabular}
    }
    \label{tab:wga}
\end{table}

\vspace{-3mm}

\subsection{Ablations}
\label{ablations}
\begin{figure}[tb]
  \centering
  \begin{subfigure}[t]{0.43\textwidth}
\includegraphics[width=\textwidth]{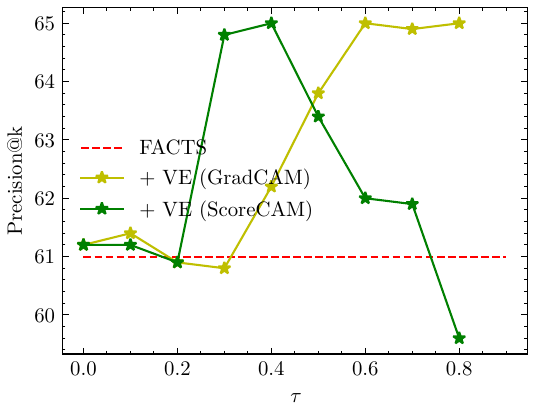}
    \caption{We plot the value of the \texttt{Precision@k} obtained after fitting the mixture model in FACTS with and without incorporating visual explanations against the threshold $\tau$, and compare ViG-FACTS (GradCAM) and ViG-FACTS (ScoreCAM) for visual explanation capabilities on the CelebA datasets.}
    \label{fig:tau_precision}
  \end{subfigure}
  ~~~~
    \begin{subfigure}[t]{0.43\textwidth}
    \includegraphics[width=\textwidth]{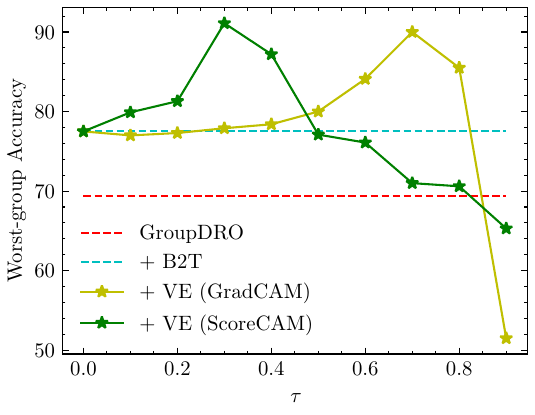}
    \vspace{-3mm}
    \caption{We plot the value of the worst-group accuracy obtained using GroupDRO with ViG-B2T groups classifier, as the threshold $\tau$ increases for GroupDRO with ViG-B2T (GradCAM) and GroupDRO with ViG-B2T (ScoreCAM) on the CelebA dataset.}
    \label{fig:tau_wga_gradcam_compressed}
  \end{subfigure}%
  \caption{Ablating the GradCAM \cite{Selvaraju_2019} mask threshold $\tau$ on the CelebA dataset for bias discovery (a) and mitigation (b). The main hyperparameter that is important for our model is the parameter $\tau$ that integrates the optimal threshold for the mask. We optimize this parameter depending on the dataset, however, our method performs better than the baseline in almost all the different choices of $\tau$. Empirically, as a rule of thumb, setting $\tau=0.5$ is a good starting point.}
  \label{fig:tau}
\end{figure}

We considered an alternative VE method (ScoreCAM \cite{wang2020scorecam}, which differently from GradCAM, does not rely on gradients to construct the class activation maps) and different thresholds $\tau$ for the generation of the binary masks. \Cref{fig:tau} shows results both for bias discovery (\Cref{fig:tau_precision}) and mitigation (\Cref{fig:tau_wga_gradcam_compressed}). We aim to assess the stability of the visual explanation methods across different threshold levels. 
We find that both, GradCAM and ScoreCAM achieve approximately similar worst group accuracies for different values of $\tau$. For instance, ScoreCAM  gets good worst-group accuracy at lower threshold values, whereas GradCAM \cite{Selvaraju_2019} requires increasing threshold values. Nevertheless, the worst group accuracy remains approximately constant for both methods. Finally, some qualititave results from these visualizations are provided in Figure \ref{fig:facts_method_with_visual_explanations} for different attention methods. Additional results are also included in Supplementary Material C.



~~~
\begin{figure}[tbh!]
    \centering
    \includegraphics[width=0.7\textwidth]{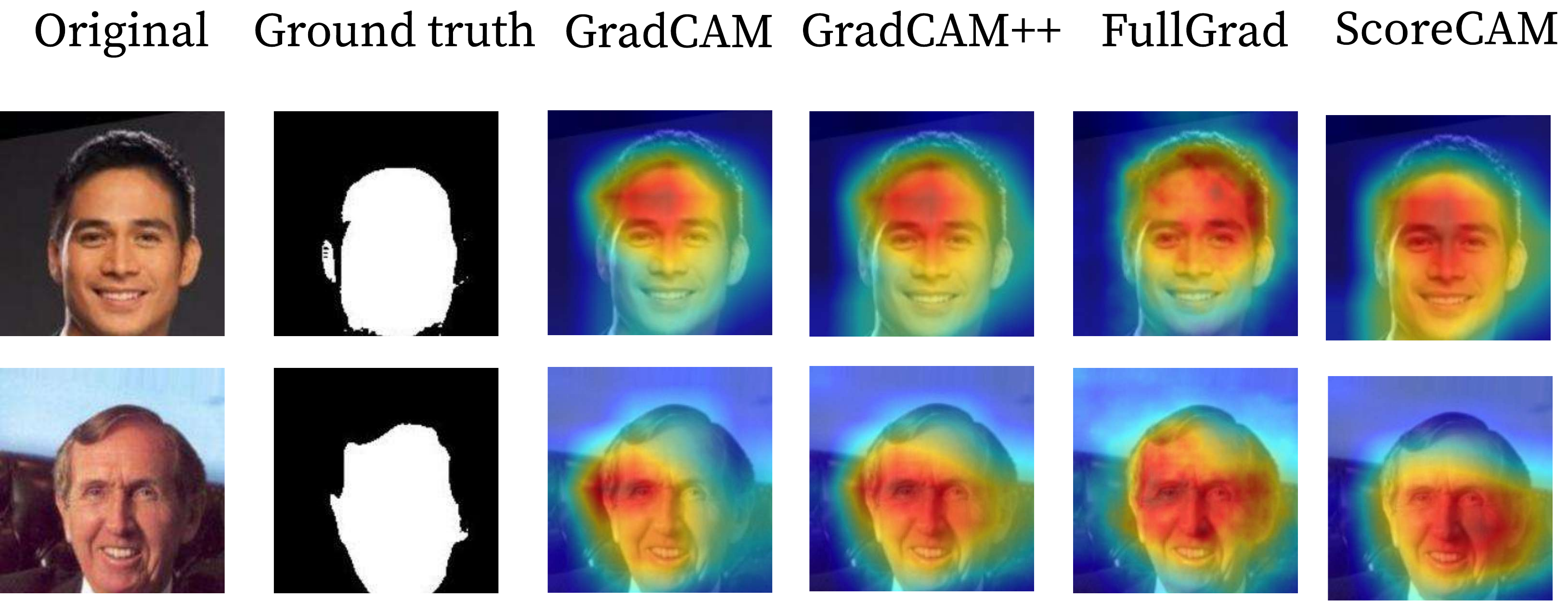}
    \caption{Visual explanations of other CAM-variants (GradCAM \cite{Selvaraju_2019}, GradCAM++ \cite{Chattopadhay_2018}, Fullgrad     \cite{srinivas2019fullgradient} and ScoreCAM \cite{wang2020scorecam}) for a biased model trained on CelebA. Compared to other CAM methods, ScoreCAM tends to better highlight the region related to the spurious feature (in this case, instead of focusing on the hair, it points to the face), for images from groups where the spurious correlation holds.}
    \label{fig:facts_method_with_visual_explanations}
\end{figure}





\vspace{-0.5cm}
\section{Conclusion}
In this paper, we introduced Visually Grounded Bias Discovery (ViG-Bias), a novel method that leverages visual explanations to uncover and address biases in visual recognition models. By integrating multimodal embeddings with visual explanation techniques, ViG-Bias not only identifies biases with higher precision but also provides insights into the nature of these biases. 
Our approach represents a significant step forward 
offering a method that is both effective and interpretable, while easily adaptable to multiple existing frameworks. However, it is not free from limitations. The type of spurious correlation that can be discovered by ViG-Bias is limited to those that can be attributed by VEs. For example, spurious correlations not related to localized artifacts or objects, like global intensity distribution shifts due to noise or different acquisition devices, will not be detected by ViG-Bias. As for Societal Impact, we believe it is overall positive as our method improves bias detection and mitigation, ultimately contributing to more fair AI systems.
As future work, we aim to investigate the integration of ViG-Bias with other bias detection and mitigation frameworks to further enhance its effectiveness and adaptability.



\section*{Acknowledgments}
We gratefully acknowledge the DATAIA program for supporting EF as a visiting professor at Université Paris-Saclay, NVIDIA Corporation for provinding GPU computing, the support of Agencia Nacional de Promoción de la Investigación, el Desarrollo Tecnológico y la Innovación (Argentina), the Google Award for Inclusion Research (AIR) Program and the SticAmSud program. This work has been partially supported by project MIS 5154714 of the National Recovery and Resilience Plan Greece 2.0 funded by the European Union under the NextGenerationEU Program as well as the ANR Hagnodice ANR-21-CE45-0007.

%
%
\bibliographystyle{splncs04}
\bibliography{egbib}


\end{document}